\documentclass[10pt,twocolumn,letterpaper]{article}

\usepackage[pagenumbers]{cvpr} 

\usepackage{graphicx}
\usepackage{amsmath}
\usepackage{amssymb}
\usepackage{booktabs}

\usepackage[pagebackref,breaklinks,colorlinks]{hyperref}

\usepackage[capitalize]{cleveref}
\crefname{section}{Sec.}{Secs.}
\Crefname{section}{Section}{Sections}
\Crefname{table}{Table}{Tables}
\crefname{table}{Tab.}{Tabs.}
\newcommand{\cunet}{CU-Net\xspace}

\def\cvprPaperID{9637} 
\def\confName{CVPR}
\def\confYear{2023}

\begin{document}

\title{CU-Net: Real-Time High-Fidelity Color Upsampling for Point Clouds}

\author{Lingdong Wang\\
University of Massachusetts Amherst\\
\and
Mohammad Hajiesmaili\\
University of Massachusetts Amherst\\
\and
Jacob Chakareski\\
New Jersey Institute of Technology\\
\and
Ramesh K. Sitaraman\\
University of Massachusetts Amherst\\
}
\maketitle

\begin{abstract}
Point cloud upsampling is essential for high-quality augmented reality, virtual reality, and telepresence applications, due to the capture, processing, and communication limitations of existing technologies. Although geometry upsampling to densify a point cloud's coordinates has been well studied, the upsampling of the color attributes has been largely overlooked. In this paper, we propose CU-Net, the first deep-learning point cloud color upsampling model that enables low latency and high visual fidelity operation. CU-Net achieves linear time and space complexity by leveraging a feature extractor based on sparse convolution and a color prediction module based on neural implicit function. Therefore, CU-Net is theoretically guaranteed to be more efficient than most existing methods with quadratic complexity. Experimental results demonstrate that CU-Net can colorize a photo-realistic point cloud with nearly a million points in real time, while having notably better visual performance than baselines. Besides, CU-Net can adapt to arbitrary upsampling ratios and unseen objects without retraining. Our source code is available at https://github.com/UMass-LIDS/cunet.
\end{abstract}

\section{Introduction}

The point cloud is a fundamental 3D representation. Dense point clouds are required for the high-quality operation of many applications like rendering, surface reconstruction, and semantic understanding in augmented/virtual reality (AR/VR) \cite{yuzu, relightable, ar} as well as in telepresence \cite{telepresence, holoportation, virtualcube, avatar}. Unfortunately, hardware, storage, and computational constraints in capturing, compressing, and processing point clouds generally lead to sparse instances. Similarly, existing networking capabilities prevent the transmission of dense point clouds due to their overwhelming data volume. Therefore, a point cloud upsampling technique is essential for the densification of point clouds in order to enable high visual fidelity.

Existing studies \cite{punet, 3pu, pugan, dispu, pugcn, pugacnet, putransformer, pudense, metapu, deepmag, neuralpoints} have developed point cloud \textbf{geometry upsampling} to increase the number of spatial point coordinates, but have largely ignored the color attributes associated with the points. We argue that {\bf an efficient and scalable color upsampling} is critical in many telepresence scenarios. For example, during 3D video conferencing or remote collaboration, a human avatar represented as a colored point cloud with millions of points needs to be streamed in real time. This is presently infeasible to be accomplished using state-of-the-art point cloud compression standards \cite{mpegcomp} and available network bandwidth, thereby limiting the scope or quality of related applications. 

As an alternative approach, we advocate streaming a degraded low-quality point cloud and then upsampling it as post-processing at the destination. The colors of a human avatar must be carefully processed during upsampling, as artifacts or blur on the human face can lead to the uncanny valley effect and severely harm the visual quality. This motivates us to develop a color upsampling method that can process millions of points in real time with satisfying perceptual performance.

\begin{figure}[ht]
  \centering
  \includegraphics[width=0.9\linewidth]{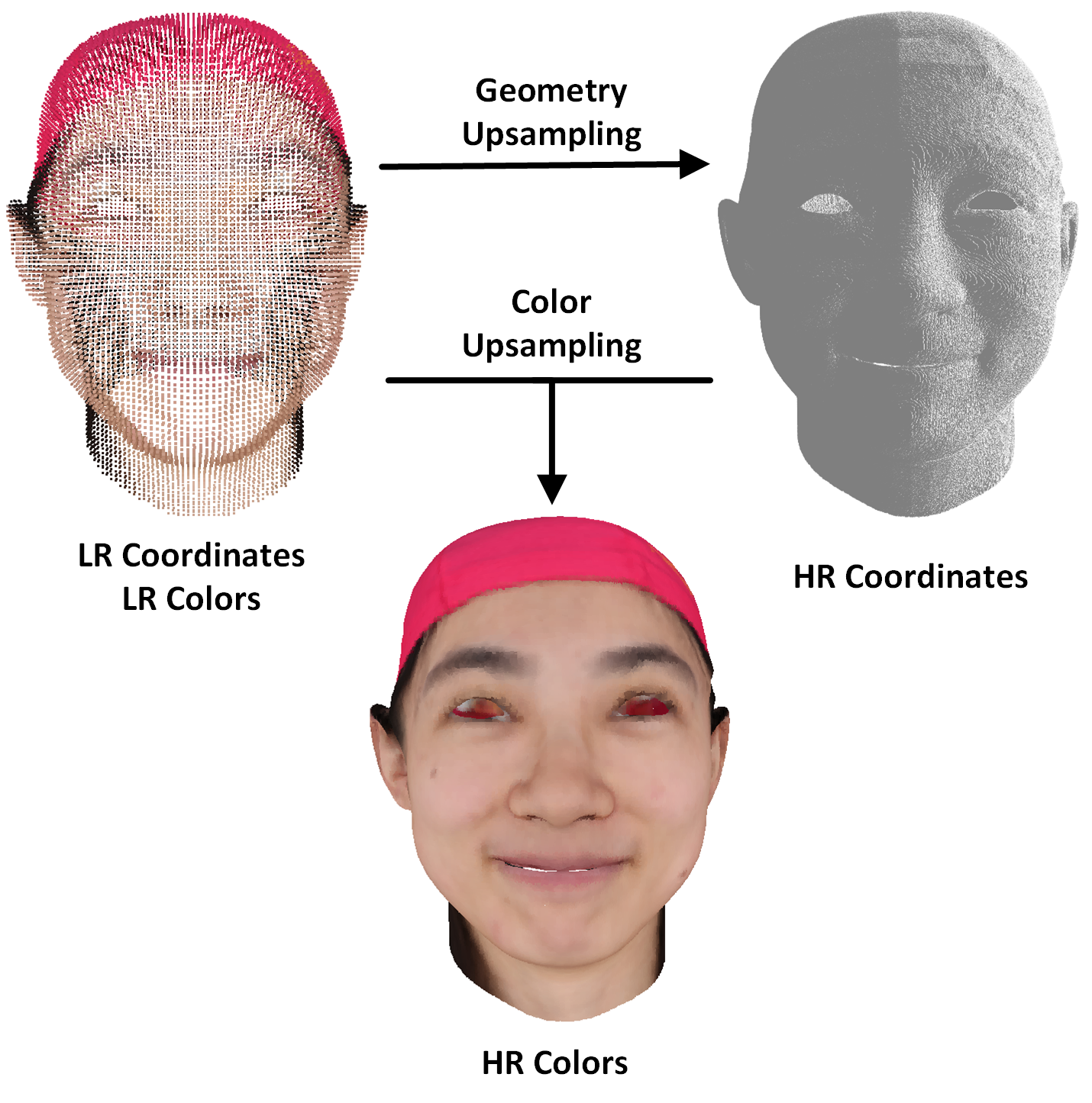}
   \caption{Point cloud geometry upsampling and color upsampling.}
   \label{fig:task}
\end{figure}

\begin{figure*}[ht]
  \centering
   \includegraphics[width=\linewidth]{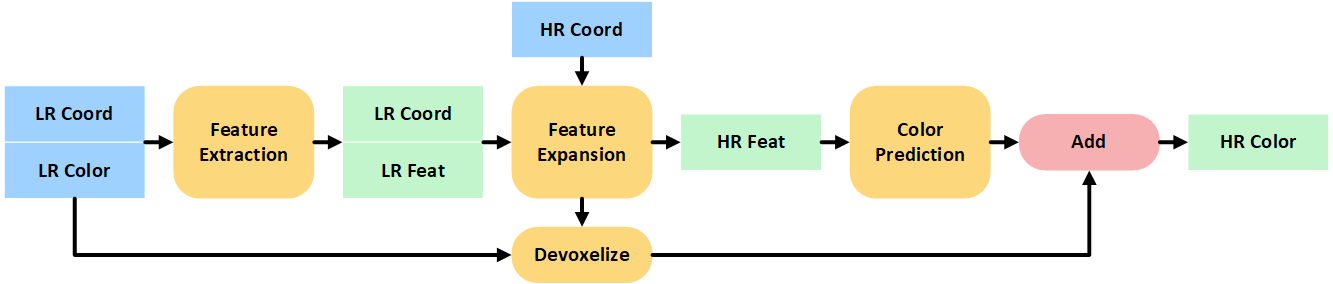}
   \caption{Workflow of CU-Net.}
   \label{fig:flow}
\end{figure*}


In this work, we focus on point cloud color upsampling, which could be viewed as a generalization of super resolution from 2D to 3D. As shown in \cref{fig:task}, color upsampling can be applied after an existing geometry upsampling method. The inputs to color upsampling are a low-resolution (LR) sparse point cloud with both point coordinates and colors, as well as a high-resolution (HR) dense point cloud comprising only spatial coordinates. The output is a set of color labels for the HR point cloud. 

Efficiency is the major challenge in achieving real-time processing of large-scale point clouds. Prior color upsampling methods are based on interpolation \cite{yuzu, fractionalsr} or optimization \cite{freqselect, fgtv} that are computationally inefficient to implement. Deep-learning methods, though they merely process coordinates for geometry upsampling, can only work with a small-scale point cloud having thousands of points. When applied to a large-scale cloud with millions of points, these methods have to follow a divide-and-merge pattern and consume tens of minutes due to their poor memory and runtime efficiency.  Instead, we introduce \cunet, which to the best of our knowledge, is the first deep-learning point cloud color upsampling method that enables low latency and high visual quality operation.


The inefficiency of existing upsampling methods is rooted in their $O(N_h^2)$ complexity, where $N_h$ is the size of the HR point cloud. Such quadratic complexity usually comes from the deep-learning feature extractors they adopt, like PointNet++ \cite{pointnet2}, Dynamic Graph CNN (DGCNN) \cite{dgcnn}, Graph Convolutional Network (GCN) \cite{pugcn, pugacnet}, or Transformer \cite{pct, pttrans}. We comprehensively analyze the complexity of existing methods in \cref{sec:deep}.

To pursue low complexity, \cunet utilizes four efficient modules: feature extraction, feature expansion, color prediction, and devoxelization. The workflow of \cunet is illustrated in \cref{fig:flow}. First, the feature extraction module extracts deep-learning color features for the LR point cloud, namely LR features. Then, the feature expansion module expands the LR features towards the HR point cloud to be HR features. Next, the color prediction module transforms HR features into residual colors. Finally, the devoxelization module generates coarse colors, which are added to the previously learned residuals to be the final HR colors. 

We carefully design each module of \cunet to achieve efficiency and scalability. The feature extraction module is based on sparse convolution \cite{minkowski, torchsparse}, the color prediction module forms a neural implicit function (neural field) \cite{nerf, neuralpoints}, and the other two modules are composed of simple tensor operations. After all, \cunet achieves $O(N_h)$ time and space complexity, which is theoretically guaranteed to be more efficient than most prior works about color upsampling or geometry upsampling. A detailed explanation of our method's complexity is provided in \cref{sec:complex}. Besides, the neural implicit function design allows \cunet to perform arbitrary-ratio upsampling without retraining.


We summarize our contributions as follows:

\begin{enumerate}
    \item We propose \cunet, the first deep-learning model for point cloud color upsampling featuring low complexity, low latency, and high visual quality performance. 
    \item \cunet is efficient in terms of runtime, memory consumption, and storage space. With $O(N_h)$ time and space complexity, \cunet can colorize a photo-realistic point cloud with nearly a million points within 30ms latency using a 3MB model. This can fit within an overall frame rate of 30fps for point cloud videos, suitable for real-time deployment.
    \item \cunet can adapt to an arbitrary upsampling ratio once trained properly, which avoids retraining and storing multiple models.
    \item Through comprehensive experiments with several reference methods, we verify that \cunet significantly outperforms the latter in visual quality both quantitatively and qualitatively. \cunet bests other alternatives by 0.7-1.5dB in PSNR when generalized to unseen objects.
\end{enumerate}

\section{Related Works}

\subsection{Deep Learning for Point Cloud}
\label{sec:deep}
Deep learning feature extractors are widely adopted for 3D point cloud processing. The most popular five are PointNet++ \cite{pointnet2}, Dynamic Graph CNN (DGCNN) \cite{dgcnn}, Graph Convolutional Network (GCN) \cite{pugcn, pugacnet}, Transformer \cite{pct, pttrans}, and sparse-convolution-based model\cite{minkowski, torchsparse}. 

The first four methods all have $O(N^2)$ complexity. PointNet++ recursively performs ball query around all points and then applies PointNet \cite{pointnet} to learn hierarchical features. DGCNN computes K-Nearest Neighbor (KNN) multiple times to construct dynamic graphs among features, and then aggregate features via edge convolution. Both ball query and KNN are implemented as $O(N^2)$ on GPU for parallelism. GCN transforms the point cloud into a KNN graph and then performs graph convolution. Graph convolution is $O(N^2)$ as it consumes adjacent matrices. Transformer employs $O(N^2)$ self-attention mechanism to capture relationships among all points.

Sparse convolution is the only feature extractor with $O(N)$ complexity. It resembles traditional 3D convolution but exploits the sparsity of point clouds. Specifically, it stores the quantized positions of all points via a hash map and only visits occupied positions. MinkowskiEngine \cite{minkowski} is a widely-used framework for sparse convolution. While TorchSparse \cite{torchsparse} achieves 1.6$\times$ speedup towards MinkowskiEngine by balancing the irregular computation workload and reducing memory footprint. For this reason, we implement our feature extractor using TorchSparse.

\subsection{Point Cloud Geometry Upsampling}



Point cloud geometry upsampling is formulated as an optimization problem in pioneering works \cite{surface, paramfree, consolidation, ear}, but they are recently outperformed by deep-learning methods. PU-Net \cite{punet} is the first deep-learning geometry upsampling model. It extracts features with PointNet++ \cite{pointnet2} and expands the point cloud using multi-branch Multi-Layer Perceptron (MLP). 3PU \cite{3pu} adopts DGCNN \cite{dgcnn} as feature extractor and upsamples point cloud in a patch-based progressive way. With the same feature extractor, PU-GAN \cite{pugan} proposes a Generative Adversarial Network (GAN) to upsample points. Dis-PU \cite{dispu} firstly generates a dense but coarse point cloud and then refines the locations. PU-GCN \cite{pugcn} employs graph convolution for feature extraction and a novel NodeShuffle module for upsampling. PU-GACNet \cite{pugacnet} further introduces attention mechanism into graph convolution to capture global-range relations. PU-Dense \cite{pudense} adopts a U-Net feature extractor based on sparse convolution. Due to the efficient sparse convolution, it can process large-scale point clouds much faster than other methods.


Some geometry upsampling methods support arbitrary-ratio upsampling. Meta-PU \cite{metapu} employs meta-learning to predict the weight of graph convolution kernels in adaptation to different ratios. Yue \etal \cite{deepmag} interpolates points through an affine combination of neighboring points projected onto the tangent plane and a further refinement. NeuralPoints \cite{neuralpoints} transforms a 2D square patch to fit a 3D local surface through a neural implicit function \cite{nerf}, which allows the model to sample any number of points from the surface. With a similar design, \cunet also supports arbitrary-ratio upsampling.

\subsection{Point Cloud Color Upsampling}

Existing upsampling methods considering colors can be classified into three categories: interpolation-based, optimization-based, and learning-based methods.

\textbf{Interpolation-based Methods.} YuZu \cite{yuzu} develops a volumetric video streaming system that uses Nearest Neighbor interpolation to relieve the transmission burden. WAAN \cite{fractionalsr} colorizes an LR voxel using the weighted average color of parent voxels sharing an edge with it. These methods generate blurry results with low visual quality. 

\textbf{Optimization-based Methods.} FGTV \cite{fgtv} transforms a point cloud into a KNN graph, and minimizes a weighted $L_1$ norm on adjacent colors through linear programming. FSMMR \cite{freqselect} constructs a minimum-spanning tree to project 3D points onto a 2D plane. It then interpolates colors on the 2D plane by selecting and optimizing the coefficients of frequency basis functions. These methods run on CPU with prohibitive latencies and limited perceptual qualities when applied to large-scale point clouds. In contrast, our \cunet can be accelerated by GPU to generate high-fidelity colors for massive point clouds in real time.

\textbf{Learning-based Methods.} FG-GAN \cite{multiattr} utilizes colors to improve the accuracy of geometry upsampling but it doesn't upsample the colors. FeatureNet \cite{featurenet} extends PU-Net \cite{punet} with an additional $L_2$ loss on colors to upsample coordinates and colors simultaneously. However, it doesn't offer the flexibility to upsample colors individually, hence cannot be used for the color upsampling task. Besides, FeatureNet only works on a fixed and small number of points with unsatisfying visual quality. As far as we know, our \cunet is the first deep-learning method for low-latency high-fidelity point cloud color upsampling.

Other than color upsampling, several topics investigate colors in point clouds. Point cloud colorization methods \cite{densepoint, point2color, pccn} generate colors for an uncolored point cloud, but it is highly ambiguous to colorize a point cloud without any guidance. Color upsampling colorizes point clouds under the supervision of LR colors, thus returning more reasonable results. Point cloud completion aims at completing a missing part of an object. Yuki \etal \cite{completionhead} adopts GAN to complete a haired human head model with a few thousand points. Point cloud color upsampling could be viewed as the completion of scattered points instead of a whole part.


\section{The Proposed Method: \cunet}


The workflow of our point cloud color upsampling model \cunet is illustrated in \cref{fig:flow}. \cunet accepts a low-resolution (LR) point cloud with both coordinates and colors, and a high-resolution (HR) point cloud with only coordinates. It generates colors for the HR point cloud. We assume that the LR point cloud is downsampled from the HR one using voxelization, which is commonly used in octree-based point cloud compression standards \cite{mpegcomp}.

\cunet starts with the \textbf{feature extraction} module that extracts deep-learning color features from the LR point cloud. The LR features are then expanded into HR features by the \textbf{feature expansion} module. The HR features will be transformed into color residuals of the HR point cloud by the \textbf{color prediction} module. Finally, the \textbf{devoxelization} module outputs coarse HR colors, which become the final HR colors together with the previously learned residuals. Details of these modules are introduced below. 

\begin{figure}[ht]
  \centering
   \includegraphics[width=\linewidth]{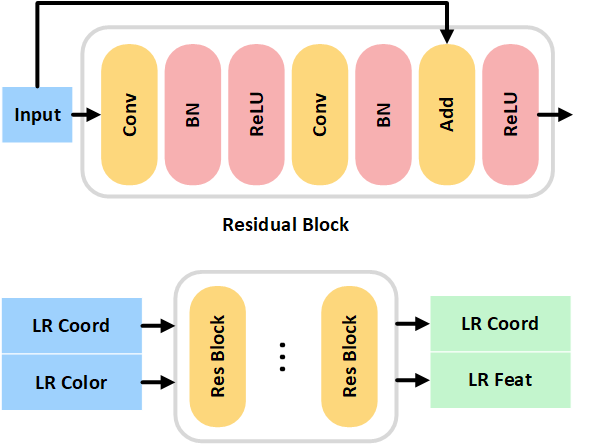}
   \caption{Structure of the feature extraction module.}
   \label{fig:feat}
\end{figure}

\subsection{Feature Extraction}

The LR point with index $i$ is defined by a 4D coordinate $p_l^i = (x_i, y_i, z_i, n_i)$ and  an RGB-format color $c_l^i = (r_i, g_i, b_i)$. Here $(x_i, y_i, z_i)$ is its position in the 3D space while $n_i$ indicates which point cloud it belongs to in this batch. This design allows us to merge point clouds of different sizes into one batch for training. The set of LR coordinates $P_l = \{p_l^i\}_i$ can be organized as a tensor with shape $(N_l, 4)$, where $N_l$ is the total number of LR points in this batch. Similarly, the set of LR colors $C_l = \{c_l^i\}_i$ is a tensor with shape $(N_l, 3)$. LR coordinates and LR colors are combined as one sparse tensor and fed into the feature extraction module. This module outputs a sparse tensor with the same LR coordinates, but deep-learning features instead of color attributes are attached to these coordinates. Formally, the point $i$ in the output sparse tensor has the coordinate $p_l^i$ and a deep-learning feature $f_l^i \in \mathbb{R}^K$, where $K$ is a hyper-parameter about the number of channels. The set of these LR features $F_l = \{f_l^i\}_i$ is represented by a tensor with shape $(N_l, K)$, and the transformation is as follows.
\begin{equation}
    P_l, F_l = \texttt{FeatureExtraction}(P_l, C_l).
    \label{eq:feat}
\end{equation}
We develop a feature extraction module based on sparse convolution. As shown in \cref{fig:feat}, the feature extractor has a simple structure with several Residual Blocks \cite{resnet}. A Residual Block is composed of convolutional layers, batch normalization layers, ReLU activation functions, and a skip connection from the input. These neural network operations are similar to the traditional ones but supported by a sparse computation backend.

\begin{figure}[ht]
  \centering
   \includegraphics[width=\linewidth]{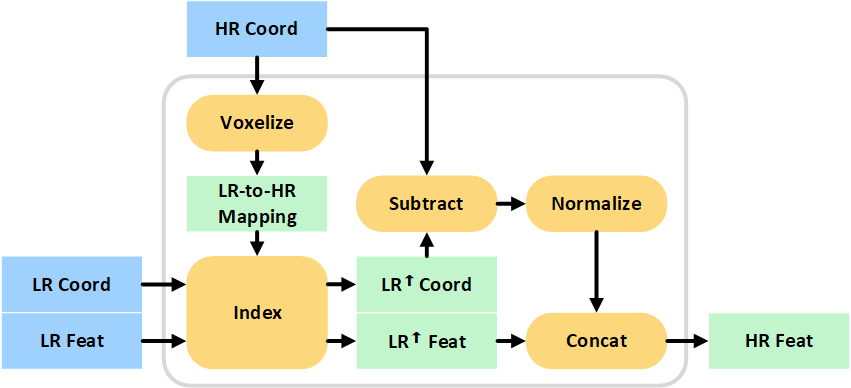}
   \caption{Structure of the feature expansion module. $\uparrow$ means that LR data is expanded to have the HR shape.}
   \label{fig:expand}
\end{figure}

\subsection{Feature Expansion}


The key to feature expansion is to compute the mapping from the LR point cloud to the HR point cloud, or the LR-to-HR mapping. Recall that we assume the LR point cloud is downsampled from HR by voxelization, which merges several HR points within the same voxel (a small cube) into one LR point at the center of this voxel. From another aspect, one LR point corresponds to several HR points that share the same coordinate after quantization. So we can trace the LR-to-HR mapping by recomputing the voxelization, and then expand the LR features using this mapping as follows.

Define the LR-to-HR mapping function as $m(\cdot)$ and $m(j) = i$ means that an HR point with index $j$ corresponds to an LR point with index $i$. LR point $i$ has the coordinate $p_l^i$ and deep feature $f_l^i$, while HR point $j$ has the coordinate $p_h^j$. The HR feature $f_h^j$ can be computed as \cref{eq:expand}.

\begin{equation}
\begin{aligned}
    f_h^j &= [f_l^{i}, \Delta p], \\
    \Delta p &= 2 \times \frac{p_h^j[0:3] - v \times p_l^i[0:3]}{v - 1} - 1, \\
    i &=m(j).
\end{aligned}
\label{eq:expand}  
\end{equation}

Here $v$ is the voxel side length and $\Delta p \in [-1, 1]$. In other words, the deep feature of an HR point is the concatenation of its corresponding LR point's feature $f_l^i$ and the normalized difference of their 3D positions $\Delta p$. The explanation for such a design is given  in \cref{sec:pred}.

We illustrate the feature expansion module in \cref{fig:expand}. The module accepts LR coordinates, LR features, and HR coordinates as inputs. It outputs the set of HR features $F_h = \{ f_h^j\}_i$ as a tensor with shape $(N_h, K+3)$, where $N_h$ is the number of HR points.

\begin{figure}[ht]
  \centering
   \includegraphics[width=\linewidth]{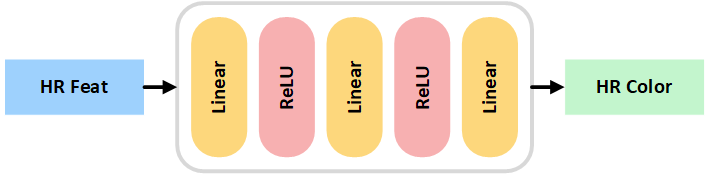}
   \caption{Structure of the color prediction module.}
   \label{fig:pred}
\end{figure}

\subsection{Color Prediction}
\label{sec:pred}


Denote the residual color predicted by this module with index $j$ as $r_h^j$. The set of residual HR colors is $R_h = \{ r_h^j \}_j$, represented by a tensor with shape $(N_h, 3)$. Then the transformation in this step is given as.
\begin{equation}
    R_h = \texttt{ColorPrediction}(F_h).
\label{eq:pred}  
\end{equation}
As shown in \cref{fig:pred}, the color prediction model is simply a 3-layer Multi Layer Perceptron (MLP). Each layer halves the number of channels, and the last layer reduces it to 3. 

The design of our color prediction module is inspired by recent works about neural implicit function (neural field) \cite{nerf, neuralpoints}. Neural implicit function is a function parameterized by a neural work, which returns a value given a query position, conditioned on a latent feature. In our case, the LR feature is used as the latent feature to describe local color distribution within a voxel, the normalized coordinate shift is used as the query, and the output value is a residual color. We do not adopt the sine-cosine positional encoding used in methods \cite{nerf, neuralpoints} because it brings no improvement.

\subsection{Devoxelization}
\label{sec:devox}


As the inverse operation of voxelization, devoxelization assigns an LR point's color to all the HR points within the same voxel. When used alone, it incurs blocky artifacts as shown in \cref{fig:visual}. But when used in \cunet, it can force the deep-learning model to spend its capacity on the residual color details and thus improving the final performance.

Devoxelization can be efficiently performed by reusing the LR-to-HR mapping computed before. Denote the coarse color generated by devoxelization with index $j$ as $d_h^j$. Devoxelization is formulated as \cref{eq:devox}. 
\begin{equation}
    d_h^j = c_l^i, ~s.t. ~ i=m(j).
\label{eq:devox}  
\end{equation}
Define the final HR color indexed by $j$ as $c_h^j$. Then as in \cref{eq:add}, the HR color is the sum of a coarse base $d_h^j$ and a deep-learning residual $r_h^j$.
\begin{equation}
    c_h^j = d_h^j + r_h^j.
\label{eq:add}  
\end{equation}

\subsection{Complexity Analysis}
\label{sec:complex}

The feature extraction module is based on sparse convolution. Sparse convolution pre-stores all quantized coordinates using a hash map and only visits occupied positions. Thus it has an average complexity of $O(N_l)$. The feature extraction module only comprises linear-complexity operations, thus the complexity of feature expansion is $O(N_h)$. Note that the voxelization operation can be decomposed into one quantization operation over HR coordinates, and one \textit{unique} operation on the quantized coordinates to remove duplicated positions. The color prediction module is an MLP whose complexity is $O(N_h)$. And the devoxelization is an indexing operation with $O(N_h)$ complexity. 

In summary, \cunet has the complexity of $O(N_l + N_h) = O(N_h)$ for both time and space. Experimental proof of this linear complexity can be found in \cref{sec:eff}. Although methods like KNN can gain $O(N_h log (N_h))$ complexity on CPU using KD-Tree, it's hard to do so on GPU in parallel. After all, most existing methods can only achieve $O(N_h^2)$ complexity. Hence our method is significantly more efficient than them in terms of both runtime and memory.


\section{Experiments}

\subsection{Datasets}
We use two datasets in our experiments. 

\textbf{The FaceScape Dataset.} We train \cunet on the FaceScape \cite{facescape} dataset. FaceScape contains 16,940 human head models captured from 847 subjects performing 20 different expressions. We use 80\% of samples for training, 10\% for validation, and 10\% for evaluation. A FaceScape model is originally a 3D mesh with a $4096\times4096$-resolution texture map. We randomly sample $1\times10^6$ points from the surface and then voxelize them to improve uniformity. The final point clouds lie in a $1000^3$ space with around $782 \times 10^3$ points. They are further voxelized to be the ground truth and input data for three tasks: $2\times$, $5\times$, and $10\times$ upsampling. 

\textbf{The MPEG 8i Dataset.} Beyond FaceScape, we also evaluate our method on the MPEG 8i \cite{mpeg8i} dataset. It contains 6 voxelized point cloud models. We downsample them into a $1024^3$ space with an average of $848\times10^3$ points. We investigate $2\times$, $4\times$, and $8\times$ upsampling tasks on this dataset. 
More details about both datasets are given in~\cref{tab:data}. Note that the task name refers to the upsampling ratio w.r.t. the space, while the upsampling ratio w.r.t. the number of points is listed in the last column of \cref{tab:data}. 



\begin{table*}[ht]
  \caption{Numerical details of color upsampling tasks on the FaceScape \cite{facescape} and the MPEG 8i \cite{mpeg8i} datasets.}
  \vspace{-2mm}
  \centering
  \label{tab:data}
  \begin{tabular}{ccccccc}
    \toprule
    Dataset & Task & Input Space & Output Space & \#Input Points  & \#Output Points & Up Ratio \\
    \midrule
    FaceScape & $2\times$ & $100^3$ & $200^3$ & $25 \times 10^3$ & $97 \times 10^3$ & $\sim 4 \times$ \\
    & $5\times$ & $100^3$ & $500^3$ & $25 \times 10^3$ & $445 \times 10^3$ & $\sim 18 \times$ \\
    & $10\times$ & $100^3$ & $1000^3$ & $25 \times 10^3$ & $782 \times 10^3$ & $\sim 31 \times$ \\
    \midrule
    MPEG 8i & $2\times$ & $512^3$ & $1024^3$ & $236 \times 10^3$ & $848 \times 10^3$ & $\sim 4 \times$ \\
    & $4\times$ & $256^3$ & $1024^3$ & $61 \times 10^3$ & $848 \times 10^3$ & $\sim 14 \times$ \\
    & $8\times$ & $128^3$ & $1024^3$ & $15 \times 10^3$ & $848 \times 10^3$ & $\sim 56 \times$ \\
  \bottomrule
\end{tabular}
\end{table*}

\subsection{Settings}
\textbf{Implementation Details.} 
We implement \cunet using PyTorch \cite{pytorch} framework, while sparse convolution is based on TorchSparse \cite{torchsparse}. Baselines are implemented by Numpy \cite{numpy} and Scikit-learn \cite{sklearn}. All experiments run on Intel Xeon Silver 4214R CPU and NVIDIA RTX 2080ti GPU.

\textbf{Hyper-Parameters.} We optimize \cunet towards the Mean Square Error (MSE) loss function using an Adam optimizer. The learning rate is 0.001 and decays by 0.1 every 10 epochs. The model is trained for 25 epochs in total with weight decay of 0.0001. The number of residual blocks is 4.  For 2$\times$ upsampling, we set the number of channels $K$ as 32, and the batch size $B$ as 16. We use $K=64, B=8$ for 5$\times$ upsampling, and $K=64, B=4$ for 10$\times$ upsampling. 

\textbf{Metric.} 
We adopt the Peak Signal-to-Noise Ratio (PSNR) as the objective metric of visual quality. We evaluate all methods using the average PSNR over RGB channels of all points.

\subsection{Baselines}
We compare \cunet with the following baselines: devoxelization, Nearest Neighbor (NN),  K-Nearest Neighbor (KNN), WAAN \cite{fractionalsr}, FGTV\cite{fgtv}, and FSMMR\cite{freqselect}. 

Devoxelization spreads an LR point's color towards HR points in the same voxel. NN assigns a point with its nearest neighbor's color. NN is equivalent to devoxelization in our case, because an LR point at the center of a voxel is exactly the nearest neighbor of HR points within this voxel. Hence, in experimental results both are reported together. KNN uses the average color of a point's K (K=3 in our experiments) nearest neighbors. WAAN collects colors from parent voxels sharing an edge with the target voxel in a weighted average manner. To apply WAAN for an arbitrary ratio, we generalize it to reject nonadjacent neighbors using ball query. FGTV \cite{fgtv} constructs a KNN graph over the point cloud and then minimizes a weighted $L_1$ norm on colors via linear programming. FSMMR \cite{freqselect} projects 3D points onto the 2D plane through a minimum-spanning tree, and then interpolates colors on the 2D plane by optimizing the coefficients of frequency basis functions. Since FGTV and FSMMR have high complexity, we follow a divide-and-merge pattern to apply them to large-scale point clouds. For FGTV, we sample overlapped patches, run the algorithm on each patch, and then colorize the point cloud using the average colors from the patches. As for FSMMR, we divide the space into non-overlapped blocks and run the method from block to block.

\section{Result Analysis}

\begin{figure*}[ht]
\vspace{-4mm}
  \centering
   \includegraphics[width=0.9\linewidth]{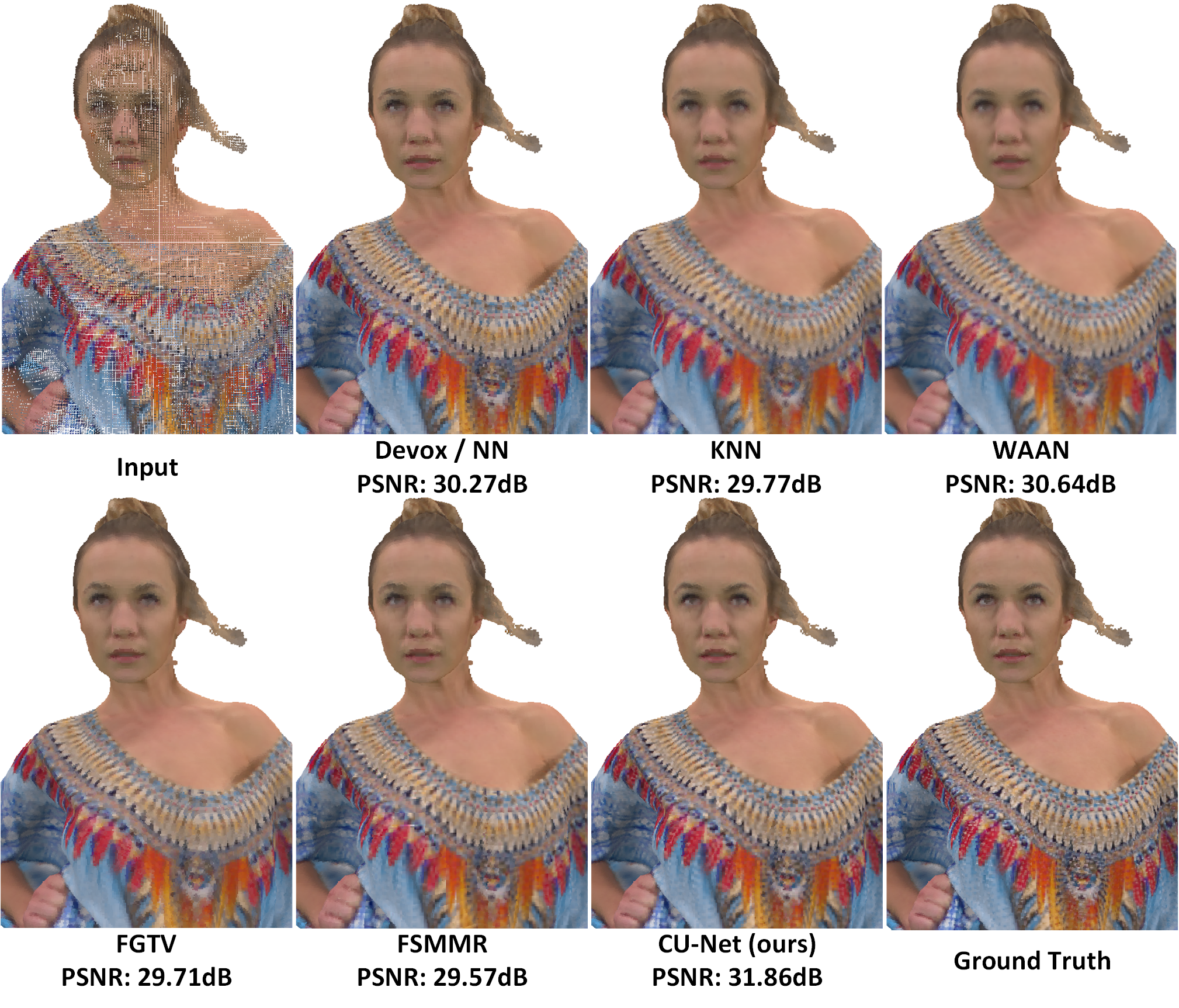}
   \caption{Results of 2$\times$ color upsampling on the MPEG 8i dataset. Zoom in for details.}
   \label{fig:visual}
\end{figure*}

\begin{table*}[ht]
  \caption{Visual qualities of color upsampling methods, measured in PSNR (dB). The best result is marked \textbf{bold}. }
  \vspace{-2mm}
  \centering
  \label{tab:psnr}
  \begin{tabular}{cccccccc}
    \toprule
     Dataset & Task & Devox/NN  & KNN & WAAN\cite{fractionalsr} & FGTV\cite{fgtv} & FSMMR\cite{freqselect}  & CU-Net(ours)  \\
    \midrule
    FaceScape& $2\times$  & 31.23 & 30.62 & 31.33 & - & - & \textbf{32.09} \\
    & $5\times$ & 30.42 & 30.49 & 30.65 & - & - & \textbf{31.32}\\
    & $10\times$ & 30.37 & 30.52 & 30.43 & - & - & \textbf{31.26}\\
    \midrule
    MPEG 8i & $2\times$  & 34.70 & 34.39 & 35.30 & 34.32 & 33.94 & \textbf{36.78} \\
    & $4\times$ & 30.28 & 30.20 & 30.48 & 30.30 & 26.07 & \textbf{31.52} \\
    & $8\times$ & 27.35 & 27.23 & 27.16 & 27.41 & 27.35 & \textbf{28.09} \\
  \bottomrule
\end{tabular}
\end{table*}

\subsection{Visual Quality}
We compare the visual qualities of color upsampling methods quantitatively in \cref{tab:psnr}. FGTV and FSMMR are only evaluated on the small-scale (w.r.t. the number of samples) dataset MPEG 8i due to their low efficiency. FSMMR has low performance, probably due to its unstable 3D-to-2D projection via minimum spanning tree. Other baseline methods show advantages in different tasks. However, \cunet significantly outperforms all baselines in all tasks by 0.7-1.5dB in PSNR. Note that in the three MPEG 8i tasks, we adopt a \cunet model originally trained for FaceScape $10\times$ upsampling task. Experimental results verify that \cunet can generalize to unseen objects with an arbitrary upsampling ratio, while still outperforming existing methods.


We illustrate representative $2\times$ upsampling results in \cref{fig:visual}. Devoxelization or NN leaves blocky artifacts that are obvious on the human face. KNN, FGTV, and FSMMR output coarse colors with low visual quality. WAAN achieves the highest PSNR score among all baselines, but the generated texture is overly-smooth. In contrast,  our \cunet learns to generate sharp and clear details, which improve the visual quality both quantitatively and qualitatively.

\begin{table*}[ht]
  \caption{Latencies of color upsampling methods on the FaceScape dataset.}
  \vspace{-2mm}
  \centering
  \label{tab:eff}
  \begin{tabular}{cccccccc}
    \toprule
    Task & Property & Devox/NN  & KNN & WAAN\cite{fractionalsr} & FGTV\cite{fgtv} & FSMMR\cite{freqselect}  & CU-Net(ours)  \\
    \midrule
    $2\times$ & CPU Latency(s) & 0.01 & 0.22 & 0.32 & 138.71 & 177.43 & 0.36 \\
    & GPU Latency(ms) & 1.73 & - & - & - & - &  13.02 \\
   \midrule
    $5\times$ & CPU Latency(s) & 0.04 & 0.96 & 1.38 & 615.18 & 901.42 & 0.72 \\
    & GPU Latency(ms) & 4.69 & - & - & - & - &  22.17 \\
    \midrule
    $10\times$ & CPU Latency(s) & 0.06 & 1.66 & 2.38 & 1101.92 & 3282.69 & 0.87 \\
    & GPU Latency(ms) & 7.39 & - & - & - & - &  29.06 \\
  \bottomrule
\end{tabular}
\end{table*}

\subsection{Efficiency}
\label{sec:eff}

We compare the end-to-end latencies, including both data movement and computation, of different methods on CPU and GPU in \cref{tab:eff}. All latencies are measured on the FaceScape dataset using one CPU or GPU. We find that devoxelization or NN has the lowest latency on both CPU and GPU. KNN and WAAN run on CPU with higher latency, and it is nontrivial to apply them for large-scale point clouds on GPU. Optimization-based methods like FGTV and FSMMR consume minutes to upsample one object because they can only run on CPU from patch to patch. \cunet is able to run on CPU but will gain a significant speedup on GPU. When deployed on GPU, \cunet achieves real-time ($>$30 FPS) inference by having 77 FPS for 2$\times$ upsampling and 45 FPS for 5$\times$ upsampling. \cunet can even achieve 34 FPS for 10$\times$ upsampling, which requires the model to predict around $782\times10^3$ colors for each object.

GPU latency of \cunet is strongly linear to the number of HR points in \cref{tab:data} with the coefficient of determination $R^2 = 0.9948$. Hence, it verifies \cunet's theoretical complexity $O(N_h)$. The CPU latency has a sub-linear relationship with the number of HR points, as the implementation of deep learning operations on CPU is different from GPU.

\cunet is memory-efficient to process millions of points within a single GPU card during the training process. As for the model size, \cunet trained for $2\times$ upsampling is 0.77MB, while \cunet for $5\times$ or $10\times$ is 3.01MB. The storage cost of \cunet is negligible for most devices. 

In summary, \cunet has satisfying efficiency in terms of runtime, memory consumption, and storage space.

\begin{figure*}[ht]
  \centering
   \includegraphics[width=0.9\linewidth]{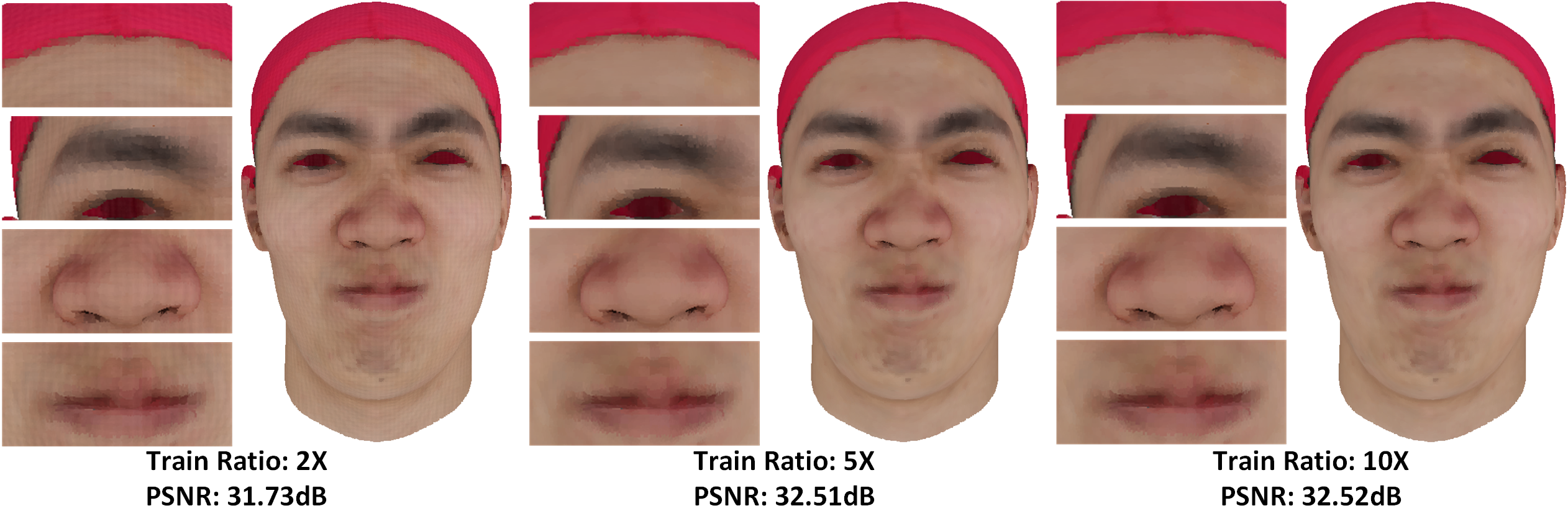}
   \caption{Arbitrary-ratio upsampling results. \cunet models trained for different upsampling ratios are used for $10\times$ upsampling.}
   \label{fig:arb}
\end{figure*}

\begin{table}[ht]
  \caption{Arbitrary-ratio upsampling results of \cunet on the FaceScape dataset, measured in PSNR (dB).}
  \vspace{-2mm}
  \centering
  \label{tab:arb}
  \begin{tabular}{c|cccc}
    \toprule
    Train  $\backslash$ Test Ratio & 2$\times$ & 5$\times$ & 10$\times$ \\
    \midrule
    2$\times$ & 32.09 & 30.65 & 30.61 \\
    5$\times$ & \textbf{32.12} & \textbf{31.32} & 31.25\\
    10$\times$ & 32.10 & \textbf{31.32} & \textbf{31.26} \\
  \bottomrule
\end{tabular}
\end{table}

\subsection{Arbitrary-Ratio Upsampling}

As discussed in \cref{sec:pred}, the color prediction module of \cunet is a neural implicit function, which enables \cunet to accept and predict colors for any number of query positions. Thus \cunet can support an arbitrary upsampling ratio and has the potential to generalize from one ratio to another after training.

To verify this property, we train \cunet towards one upsampling ratio (train ratio) and evaluates it on another ratio (test ratio). According to \cref{tab:arb}, \cunet trained for 5$\times$ upsampling has satisfying performance on a lower ratio like 2$\times$ or a higher ratio like 10$\times$. The model trained for 10$\times$ can also adapt to any ratio. However, as shown in \cref{fig:arb}, the model trained for $2\times$ upsampling doesn't generalize well to other ratios. We argue that \cunet is trivialized during $2\times$ upsampling because it only learns to expand a feature towards $2^3=8$ positions. In contrast, \cunet trained for higher ratios needs to spread a feature towards $5^3=125$ or $10^3=1000$ positions. This forces the model to learn an approximately continuous color field where the model can sample from any position.

In a nutshell, \cunet trained for a high upsampling ratio can be used for an arbitrary ratio without performance loss.


\section{Conclusion}

Point cloud color upsampling is demanded but long ignored in AR/VR and telepresence scenarios. In this paper, we present \cunet, the first deep-learning color upsampling model that enables low-latency high-fidelity operation. \cunet adopts a feature extraction module based on sparse convolution, an efficient feature expansion module, a color prediction module analogous to neural implicit function, and a devoxelization module for coarse colors. \cunet achieves $O(N_h)$ time and space complexity, which is theoretically better than most existing methods with quadratic complexity. Through extensive experiments, we show that \cunet outperforms baseline methods in visual quality while having satisfying runtime, memory, and storage efficiency. Besides, CU-Net can generalize to arbitrary upsampling ratios and unseen objects without retraining.

{\small
\bibliographystyle{ieee_fullname}
\bibliography{egbib}
}

\end{document}